\pdfoutput=1
\documentclass{article}
\usepackage[T1]{fontenc}
\usepackage{styles/iclr2027_conference,times}
\usepackage{amsmath}
\usepackage{array}
\usepackage{booktabs}
\usepackage{graphicx}
\usepackage{xcolor}
\usepackage{tikz}
\usepackage{pgfplots}
\usepackage{placeins}
\usepackage{float}
\usepackage{rotating}
\usepackage[hidelinks]{hyperref}
\usepackage{url}
\usepgfplotslibrary{colormaps}
\pgfplotsset{compat=1.18}

\title{Temporal-Attention Head Specialization During Video Diffusion Training}
\author{Taewoo Ha$^{1}$, Shafayat Mowla Anik$^{2}$, Dae Yeol Lee$^{3}$, Byeong Kil Lee$^{2}$, Jeeho Ryoo$^{4}$ \\
$^{1}$Independent Researcher, $^{2}$University of Colorado - Colorado Springs \\
$^{3}$Dolby Laboratories, $^{4}$Fairleigh Dickinson University \\
\texttt{txh2120@gmail.com, sanik@uccs.edu, DaeYeol.Lee@dolby.com} \\
\texttt{blee@uccs.edu, j.ryoo@fdu.edu}}

\iclrfinalcopy

\begin{document}
\maketitle
\lhead{}

\begin{abstract}
Video diffusion transformers depend on temporal attention to coordinate
information across frames, yet existing analyses of this mechanism examine
only trained models. Such final-checkpoint analyses show what temporal
structure exists but not when or where it forms \emph{during} training, which
is the information needed to monitor training and to target head-level
interventions such as pruning. Population averages can also hide this
structure, because the increase in a few heads and the decrease in most other
heads offset each other in the mean. We therefore conduct a
checkpoint-resolved census of every temporal-attention head across nine
Open-Sora STDiT training runs spanning three model scales (306M to 1.03B
parameters). Each head is scored by its cross-frame attention concentration
(CFAC), the entropy-normalized concentration of its attention over the 16
frames. CFAC is 0 when a head spreads attention evenly across all frames and
approaches 1 when it routes each frame's attention to a few specific frames.
Heads are then selected under a preregistered change-point and effect-size
rule. Per-head analysis shows that a small minority of heads, roughly
4--13\% in full-grid runs, becomes strongly concentrated during training,
while aggregate CFAC stays flat or decreases in every run. Across seeds,
selected heads repeatedly appear in the first temporal block, but the
specific heads selected within that block differ across seeds. In the 760M
runs, selected heads show two main recurring frame-to-frame attention
patterns. In the first, a self-frame diagonal, each frame attends mostly to
itself. In the second, an adjacent-frame band, each frame attends mostly to
its neighbors. These patterns recur across runs even when they emerge in
different heads (i.e., at different layer and head indices). Our correlation
and ablation experiments do not establish that these heads affect generated
video quality. The census nevertheless identifies when and where temporal
specialization forms during training. The same checkpoint-resolved, per-head
analysis under fixed selection rules can provide this information for other
factorized video diffusion transformers and, with adapted routing metrics,
for joint spatio-temporal architectures.
\end{abstract}

\section{Introduction}

Video diffusion models combine scalable denoising objectives with transformer
backbones and flexible spatial--temporal computation
~\citep{ho2020ddpm,rombach2022latent,peebles2023dit,ho2022video}. In
factorized architectures, spatial attention processes tokens within a frame
while temporal attention exchanges information across frames at corresponding
spatial locations~\citep{bertasius2021timesformer,arnab2021vivit,
zheng2024opensora}. Because temporal attention is the only operation in such
a model that exchanges information across frames, its heads are the natural
unit for studying how temporal structure develops.

Existing analyses of video diffusion attention cover recurrent patterns,
sparsity, sinks, temporal correspondence, and the functional split between
spatial and temporal modules~\citep{wen2025attention,xi2025sparse,
liu2025understanding,nam2025difftrack}, but they examine final or nearly
trained checkpoints, leaving unclear how temporal-attention structure
develops during training. Studying this process is important for two reasons.
First, specialization may occur in only a small subset of heads and therefore
be obscured by averages over all heads. Second, identifying when and where
such specialization emerges may provide useful signals for training
diagnostics, pruning, and studying its relationship to generation quality.

\begin{figure}[t]
\centering
\includegraphics[width=0.85\linewidth]{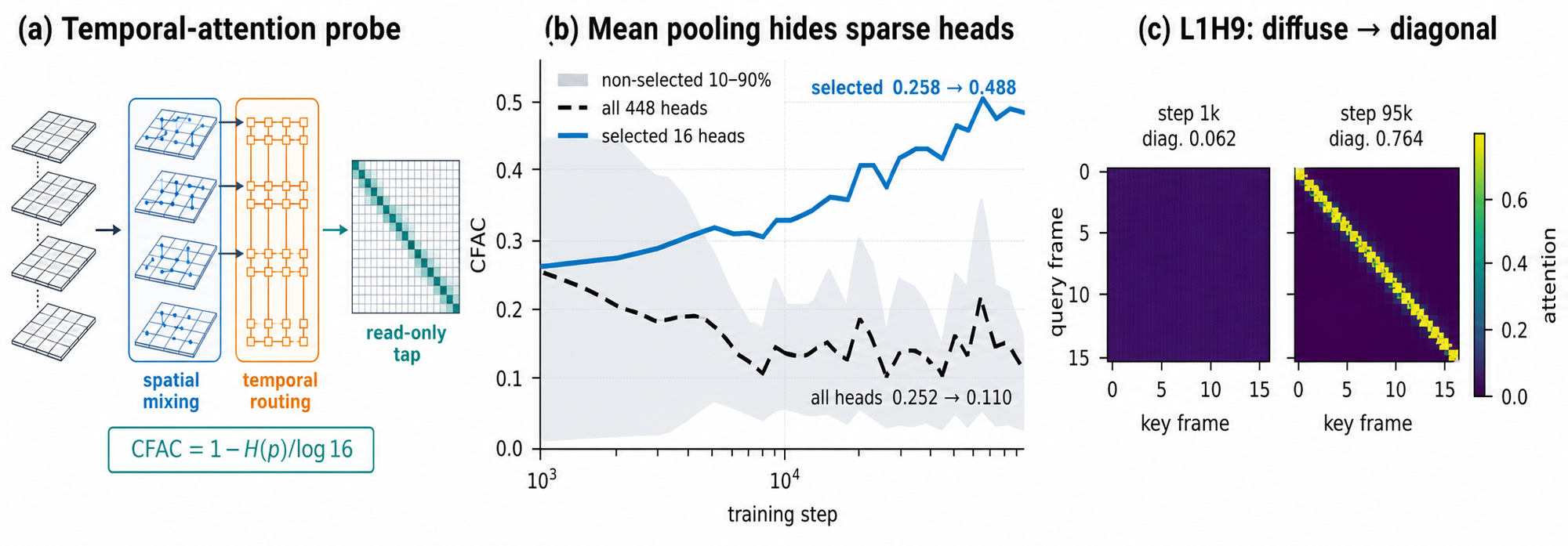}
\caption{\textbf{Sparse temporal-attention specialization is not detectable
in the population mean.} (a) Factorized STDiT and the read-only temporal
probe, with CFAC summarizing each head's cross-frame attention
concentration.
(b) Population-mean CFAC is flat or decreasing while a small selected
minority rises sharply. (c) A representative selected head transitions
from uniform attention to a self-frame diagonal over training.}
\label{fig:overview}
\end{figure}

Studying how temporal attention develops during training requires an
architecture in which temporal interactions can be measured directly and a
training pipeline that can be rerun with frequent checkpointing. Open-Sora
STDiT~\citep{zheng2024opensora} satisfies both requirements. Its factorized
spatial--temporal design gives each temporal-attention head a well-defined
attention distribution over key frames, so each head can be tracked
individually during training. Its open training pipeline also allows us to
train multiple runs under matched configurations and analyze their
trajectories across checkpoints. The same analysis can be applied directly to
other factorized video diffusion transformers, such as Latte and
VDT~\citep{ma2024latte,lu2023vdt}. For joint spatio-temporal architectures
such as Wan, LTX-Video, CogVideoX, and
HunyuanVideo~\citep{wan2025wan,hacohen2024ltxvideo,yang2024cogvideox,
kong2024hunyuanvideo}, an additional aggregation step is required to convert
joint attention over spatial--temporal tokens into a frame-level attention
distribution.

Figure~\ref{fig:overview} summarizes the central finding. Panel~(a)
illustrates the measurement setup, in which a read-only probe attached to
each STDiT block records every temporal head's $16\times16$ query--key
frame attention distribution and summarizes it with cross-frame attention
concentration (CFAC, defined in Section~\ref{sec:method}). Panel~(b) plots
this quantity over training for 760M Run A. The mean over all heads
(dashed) decreases from 0.252 to 0.110, whereas the sixteen heads later
identified by our census (blue) increase from 0.258 to 0.488. The population
mean therefore does not reflect the increase in the selected heads.
Panel~(c) shows the attention map of one such head, which is
indistinguishable from uniform at step 1k and concentrates 0.764 of its mass
on the self-frame diagonal by step 95k.

These findings hold across scales, but they do not support emergence at
specific (layer, head) coordinates. Aggregate CFAC shows no clear transition.
The selected subset is sparse in every full-grid run, though its size varies
across seeds. Reproduction across runs is concentrated in the first temporal
block (L0). In every seed, L0 contains selected heads and the
population-level rank statistics reproduce. The specific heads selected
within L0 differ across seeds. Finally, in the analyzed 760M subset, selected
heads develop recurring self-frame diagonals and adjacent-frame bands even
when coordinates and directions differ across runs. Our contributions are as
follows.
\begin{itemize}
\item a checkpoint-resolved, per-head analysis of temporal-attention
development across nine STDiT training runs and three model scales, using
fixed selection criteria and cross-seed reproduction tests.
\item evidence that specialization occurs in only a small subset of
temporal-attention heads and can therefore be obscured by population-level
averages. Across runs, this specialization is reproducible primarily at the
block level rather than at specific head indices. Selected heads develop
recurring frame-to-frame attention patterns, and negative controls separate
this description from claims about generation quality or causality.
\item an analysis framework consisting of the attention-concentration metric,
the head-selection procedure, and the cross-seed localization tests, which
applies directly to other factorized video diffusion transformers and extends
to joint spatio-temporal architectures with an adapted frame-level routing
metric.
\end{itemize}

\section{Related Work}
\label{sec:related}

Diffusion models, latent diffusion, and diffusion transformers underpin
modern video systems~\citep{ho2020ddpm,rombach2022latent,peebles2023dit},
with video architectures adding temporal operators or factorized
spatial--temporal attention~\citep{ho2022video,singer2023makeavideo,
blattmann2023align,bertasius2021timesformer}. Open-Sora STDiT is well suited for developmental analysis because its temporal-attention module is explicitly separated and its training pipeline is publicly available~\citep{zheng2024opensora}. VDT
and Latte also share the factorized design~\citep{lu2023vdt,ma2024latte}. Prior studies have analyzed attention in trained video diffusion models from several perspectives, including common attention structures, sparsity, and attention sinks~\citep{wen2025attention}, attention patterns for efficient inference~\citep{xi2025sparse}, the functional roles of spatial and temporal modules~\citep{liu2025understanding}, and temporal correspondence across frames~\citep{nam2025difftrack}. In contrast, our work focuses on how temporal attention develops during training. We track every temporal-attention head across checkpoints using fixed selection criteria and compare the resulting patterns across random seeds.

In language transformers, a minority of heads often carries interpretable
structure while many are redundant~\citep{clark2019bert,voita2019heads,
michel2019sixteen,kovaleva2019dark}. Two lines of developmental work are
closest to ours. \citet{olsson2022induction} track checkpoints, observe
induction heads forming abruptly, and connect that formation to in-context
learning with causal head-level interventions in small models.
\citet{bietti2023birth} analyze the ``birth'' of the induction mechanism in
a simplified two-layer transformer, including weight dynamics and supporting
theory. Our study makes the opposite choice. We analyze a real, substantially
larger video diffusion model at the cost of mechanistic depth. We
characterize where attention concentrates and when, but we do not identify
the implemented algorithm or, given the null intervention reported here, a
link to capability. Complementary developmental tools include progress
measures for grokking~\citep{nanda2023grokking}, refined local learning
coefficients~\citep{wang2025rllc}, and checkpoint studies of
capability-specific heads and sink formation in 1B-class language
models~\citep{xu2026circuits}.

Attention weights expose routing but are not, by themselves, faithful causal
explanations of model output~\citep{jain2019attention,wiegreffe2019attention,
abnar2020flow}, and high concentration can reflect sink-like behavior rather
than a meaningful relation~\citep{xiao2024sinks,guo2024activedormant}. We
therefore use attention maps descriptively, screen for single-frame sinks,
and reserve causal claims for interventions. The intervention reported here
is null under the preregistered metric.

\section{Experimental Setting and Analysis}
\label{sec:method}

\subsection{Models, data, and training runs}
\label{sec:models}

We analyze three Open-Sora STDiT configurations trained on
UCF-101~\citep{soomro2012ucf101} at 16 frames and $256\times256$ resolution.
Within each STDiT transformer block, spatial attention operates over tokens
within a frame, whereas temporal attention operates across frames at
corresponding spatial locations (Figure~\ref{fig:overview}a). With 16-frame
inputs, each temporal head therefore yields a $16\times16$ query--key frame
attention matrix per spatial location.
We refer to the models by rounded size (306M, 760M, 1.03B), with 16/28/28
blocks and 16/16/21 temporal heads per block, respectively. Exact parameter
counts are given in Table~\ref{tab:models} (Appendix~\ref{app:stats}).
All three STDiT configurations (306M, 760M, and 1.03B) were trained from scratch.
Differences in architecture, checkpoint grids, and batch size prevent
interpreting cross-scale differences as pure effects of parameter count.

At each scale, Runs A, B, and C denote distinct random-seed training
trajectories. The primary 760M and 1.03B analyses use Runs A and B, and the
three-run extensions and the 306M analysis use all three runs. The 306M and 1.03B models have 60 log-spaced
checkpoints from 1k to 50k steps. The primary 760M grids have 27 checkpoints
from 1k to 95k. Their three-seed intersection has 16 checkpoints from 1k to
38k and is diagnostic because it falls below the 20-point minimum required
for a registered main-analysis verdict.

All runs use bf16, HybridAdam in AdamW mode, a learning rate of
$2\times10^{-5}$, zero weight decay, gradient clipping at 1.0, and a 50k-step
cap except for the longer 760M trajectory. The configured per-process batch
size is 8 for 306M and 760M and 4 for 1.03B. Seeds and output paths differ
within a scale, and the remaining recorded training configuration is held
fixed.

\subsection{Checkpoint probe and CFAC}

We write \texttt{L}$x$\texttt{H}$y$ for head $y$ in temporal block $x$,
using zero-based indices for both. For example, \texttt{L1H9} denotes
head 9 in the second temporal block.
The read-only probe patches \texttt{STDiTBlock.attn\_temp} and captures the
softmax temporal attention. Within temporal attention, the query--key logits
are computed in float32 and softmax is applied over the key-frame axis. For
each layer, the saved tensor has shape
$(\text{groups},\text{heads},16,16)$, where groups are the video-batch times
spatial positions and the final axes are query and key frames. The extractor
renormalizes every row, clips probabilities at $10^{-10}$, computes row
entropy, and averages over all groups and all 16 query frames. Thus, for head
$h$ at checkpoint $t$,
\begin{equation}
\mathrm{CFAC}_h(t) = 1 - \frac{1}{|G|\,N}\sum_{g\in G}\sum_{q=1}^{N}\frac{H\!\left(p_{h,g,q}^{(t)}\right)}{\log N},
\qquad
H(p)=-\sum_{i=1}^{N}p_i\log p_i,
\end{equation}
where $p_{h,g,q}^{(t)}$ is head $h$'s attention row for query frame $q$ in
group $g$ of the probe batch $G$, $H(p)$ is the Shannon entropy of a
normalized attention row $p$, $p_i$ is its weight on key frame $i$, and
$N=16$ is the number of frame positions. CFAC is 0 for uniform temporal
attention and approaches 1 as probability mass concentrates on fewer frame
positions~\citep{shannon1948}.
CFAC captures only how concentrated a head's routing across frames is, not
which frames receive the mass. A head that
attends to its own frame and a head that attends to a distant frame can have
the same CFAC, which is why the motif analysis below inspects the attention
maps directly.
The unit of analysis is the head trajectory
$\{\mathrm{CFAC}_h(t)\}_t$, not a mean pooled over heads. At each
checkpoint, rank 0 makes one probe call per layer on a training batch of 8
with the training text conditioning, no classifier-free guidance, and a
randomly drawn diffusion timestep, while a fixed-timestep re-probe uses
$\tau\in\{50,250,500,750,950\}$.
Spatial probes capture \texttt{STDiTBlock.attn} over $S=256$ tokens per
frame, using $1-\langle H(p)\rangle/\log S$ per head; averaging and onset
selection are detailed with Table~\ref{tab:onset} (Appendix~\ref{app:stats}).

\subsection{Operational birth census}

\begin{figure}[t]
\centering
\includegraphics[width=\linewidth]{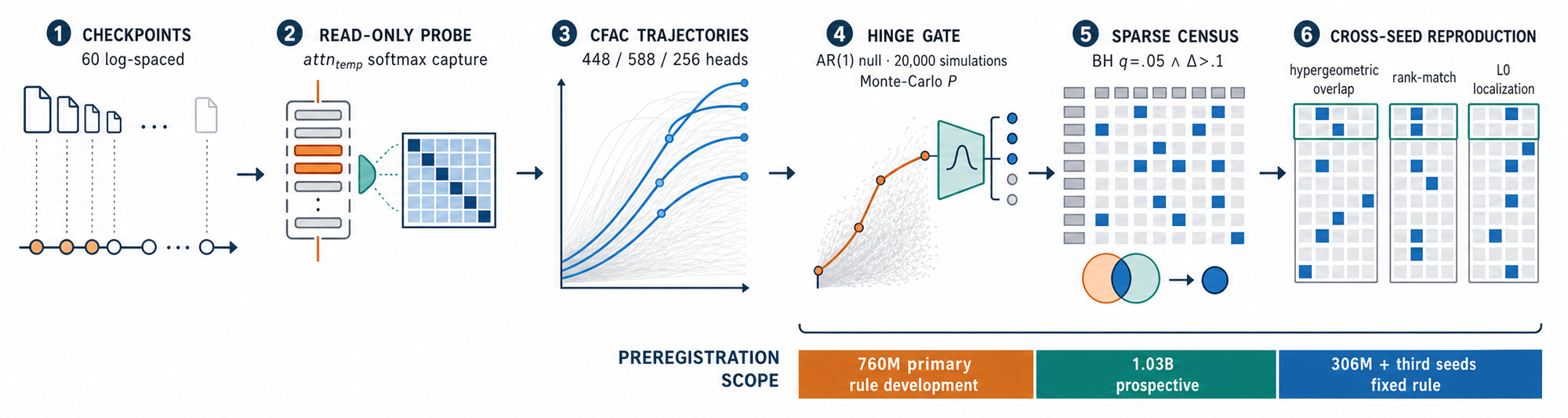}
\caption{The preregistered census pipeline, from checkpoint probe to
cross-seed localization tests. The 760M primary runs served as rule
development. The fixed rule was then applied prospectively at 1.03B and
unchanged in the three-seed extensions.}
\label{fig:pipeline}
\end{figure}

Figure~\ref{fig:pipeline} shows the pipeline, which produces one CFAC
trajectory per head (448/588/256 at the three scales). Each head trajectory is screened with a two-segment hinge model
~\citep{muggeo2003breakpoints,truong2020changepoint}. Improvement over a
single-trend model is compared with $n_{\mathrm{null}}=20000$ simulated
trajectories from a fitted linear-trend plus AR(1) Gaussian null. The gate
reports an empirical Monte Carlo p-value, direction, effect size, and hinge
and half-maximum onset estimators. The main census rule is
\begin{equation}
\operatorname{BH}(q=0.05)\ \wedge\
\Delta_h = \mathrm{CFAC}_h(t_{\mathrm{last}})
-\mathrm{CFAC}_h(t_{\mathrm{first}})>0.1,
\end{equation}
where BH denotes Benjamini--Hochberg correction across heads within a run
~\citep{benjamini1995fdr}. We call heads satisfying this rule \emph{birth
heads}. The term follows earlier transformer training analyses
~\citep{bietti2023birth,olsson2022induction}, but we use it only as
shorthand for a detected concentration increase. It does not denote the
onset of model-level temporal capability, a causal role in quality, or a
circuit. Elsewhere we use the plainer term \emph{selected heads}. Direction-hardened and alternative effect-size
thresholds are sensitivity analyses, not substitutes for this rule. The
effect-size filter creates most of the sparsity. In the four primary
760M/1.03B runs, BH alone retains 226--496 heads, whereas the conjunction
retains 16--75 (Table~\ref{tab:robust}, Appendix~\ref{app:robust}).

The 760M primary criteria were codified after those runs had been observed
and therefore constitute rule-development evidence, not prospective
confirmation. The 1.03B application was prospective and confirmatory, and
the 306M and Run C extensions then applied the already-fixed rule
unchanged. This distinction follows the confirmatory versus exploratory role
of preregistration~\citep{nosek2018preregistration}. The anonymized
supplementary archive contains both preregistration documents, the metric
and probe implementations, and the scripts and result files from which
every reported number is recomputed, and Appendix~\ref{app:robust} records
the registration timeline.

We use this change-point gate as a screening step for selecting candidate
heads rather than as a calibrated hypothesis test. Its AR(1) null indexes
checkpoints by number even though they are log-spaced in training steps, and
step-indexed surrogate checks suggest that the gate rejects more often than
its nominal level. Our conclusions therefore rest on the effect-size
threshold, repetition across runs, the sensitivity grid, and positional
reproduction rather than on individual gate p-values.

\subsection{Cross-run reproduction and motif analysis}

We evaluate reproduction at two levels. A hypergeometric test compares the
observed overlap between two census sets with chance among all head
coordinates, and an initialization-controlled rank-match test residualizes
each run's per-head net increase on the step-1{,}000 CFAC values of both runs
and tests whether high residual ranks align (defined on the broader
$\Delta_h>0.1$ population without the BH gate, so its p-value is not directly
comparable with the overlap test). To localize the signal, we repeat the
rank-match test excluding L0 and test coordinate overlap conditional on both
runs selecting L0 heads. Frame-attention dumps from eight 760M checkpoints
provide $16\times16$ maps for blocks L0--L10, and we classify final maps into
self-frame diagonal, adjacent-frame band, or diffuse patterns and screen for
previous- or next-frame asymmetry. These are descriptive routing
measurements, not causal attributions.

\section{Results}
\label{sec:results}

\subsection{Population averages obscure sparse head trajectories}

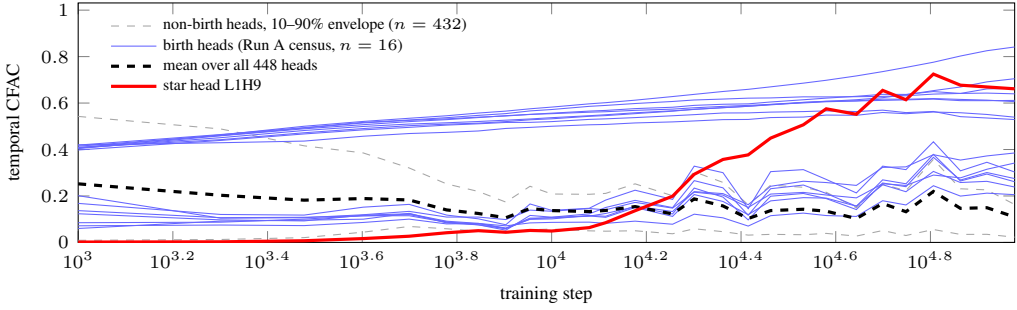
\begin{figure}[t]
\centering
\begin{tikzpicture}
\begin{axis}[
    xmode=log, width=\linewidth, height=0.34\linewidth,
    xlabel={training step}, ylabel={temporal CFAC},
    label style={font=\scriptsize}, tick label style={font=\scriptsize},
    xmin=1000, xmax=95000, ymin=0, ymax=1.031,
    ytick={0,0.2,0.4,0.6,0.8,1},
    legend pos=north west, legend cell align=left,
    legend style={font=\tiny, draw=none, fill=none, row sep=-2pt, inner sep=1pt}]
\addplot[gray!70, dashed] coordinates {(1000,0.5425) (2000,0.4885) (3000,0.4159) (4000,0.3852) (5000,0.3206) (6000,0.2519) (7000,0.2194) (8000,0.1737) (9000,0.2418) (10000,0.2084) (12000,0.2072) (13000,0.2111) (15000,0.2522) (18000,0.2010) (20000,0.3042) (23000,0.2584) (26000,0.1621) (29000,0.2465) (34000,0.2375) (38000,0.2160) (44000,0.1492) (50000,0.2504) (56000,0.2229) (64000,0.3576) (73000,0.2309) (83000,0.2263) (95000,0.1632)};
\addlegendentry{non-birth heads, 10--90\% envelope ($n=432$)}
\addplot[gray!70, dashed, forget plot] coordinates {(1000,0.0090) (2000,0.0127) (3000,0.0206) (4000,0.0442) (5000,0.0687) (6000,0.0588) (7000,0.0484) (8000,0.0404) (9000,0.0547) (10000,0.0575) (12000,0.0499) (13000,0.0484) (15000,0.0502) (18000,0.0373) (20000,0.0588) (23000,0.0463) (26000,0.0315) (29000,0.0347) (34000,0.0332) (38000,0.0389) (44000,0.0273) (50000,0.0512) (56000,0.0298) (64000,0.0556) (73000,0.0345) (83000,0.0349) (95000,0.0226)};
\addplot[blue!60, thin] coordinates {(1000,0.4199) (2000,0.4292) (3000,0.4363) (4000,0.4572) (5000,0.4688) (6000,0.4739) (7000,0.4798) (8000,0.4907) (9000,0.4952) (10000,0.5001) (12000,0.5064) (13000,0.5079) (15000,0.5144) (18000,0.5189) (20000,0.5230) (23000,0.5287) (26000,0.5294) (29000,0.5375) (34000,0.5397) (38000,0.5483) (44000,0.5473) (50000,0.5594) (56000,0.5541) (64000,0.5614) (73000,0.5583) (83000,0.5509) (95000,0.5392)};
\addlegendentry{birth heads (Run A census, $n=16$)}
\addplot[blue!60, thin, forget plot] coordinates {(1000,0.4039) (2000,0.4591) (3000,0.4891) (4000,0.5069) (5000,0.5201) (6000,0.5286) (7000,0.5376) (8000,0.5456) (9000,0.5514) (10000,0.5565) (12000,0.5654) (13000,0.5683) (15000,0.5761) (18000,0.5820) (20000,0.5900) (23000,0.5948) (26000,0.5966) (29000,0.6046) (34000,0.6142) (38000,0.6212) (44000,0.6292) (50000,0.6376) (56000,0.6371) (64000,0.6494) (73000,0.6414) (83000,0.6429) (95000,0.6397)};
\addplot[blue!60, thin, forget plot] coordinates {(1000,0.3980) (2000,0.4391) (3000,0.4670) (4000,0.4846) (5000,0.4954) (6000,0.5051) (7000,0.5107) (8000,0.5165) (9000,0.5169) (10000,0.5201) (12000,0.5297) (13000,0.5320) (15000,0.5374) (18000,0.5453) (20000,0.5501) (23000,0.5572) (26000,0.5577) (29000,0.5610) (34000,0.5633) (38000,0.5663) (44000,0.5634) (50000,0.5694) (56000,0.5589) (64000,0.5638) (73000,0.5415) (83000,0.5357) (95000,0.5290)};
\addplot[blue!60, thin, forget plot] coordinates {(1000,0.4064) (2000,0.4510) (3000,0.4818) (4000,0.4992) (5000,0.5097) (6000,0.5151) (7000,0.5206) (8000,0.5263) (9000,0.5315) (10000,0.5371) (12000,0.5455) (13000,0.5497) (15000,0.5581) (18000,0.5667) (20000,0.5725) (23000,0.5802) (26000,0.5861) (29000,0.5922) (34000,0.6025) (38000,0.6079) (44000,0.6094) (50000,0.6169) (56000,0.6163) (64000,0.6194) (73000,0.6157) (83000,0.6126) (95000,0.6061)};
\addplot[blue!60, thin, forget plot] coordinates {(1000,0.4190) (2000,0.4645) (3000,0.4942) (4000,0.5168) (5000,0.5270) (6000,0.5371) (7000,0.5470) (8000,0.5546) (9000,0.5662) (10000,0.5722) (12000,0.5858) (13000,0.5898) (15000,0.5980) (18000,0.6068) (20000,0.6132) (23000,0.6172) (26000,0.6211) (29000,0.6221) (34000,0.6268) (38000,0.6265) (44000,0.6274) (50000,0.6287) (56000,0.6314) (64000,0.6386) (73000,0.6414) (83000,0.6610) (95000,0.6632)};
\addplot[blue!60, thin, forget plot] coordinates {(1000,0.4082) (2000,0.4629) (3000,0.5001) (4000,0.5209) (5000,0.5345) (6000,0.5453) (7000,0.5570) (8000,0.5653) (9000,0.5753) (10000,0.5826) (12000,0.5966) (13000,0.6016) (15000,0.6124) (18000,0.6280) (20000,0.6374) (23000,0.6498) (26000,0.6591) (29000,0.6698) (34000,0.6857) (38000,0.6981) (44000,0.7168) (50000,0.7354) (56000,0.7535) (64000,0.7758) (73000,0.8025) (83000,0.8253) (95000,0.8414)};
\addplot[blue!60, thin, forget plot] coordinates {(1000,0.4087) (2000,0.4424) (3000,0.4721) (4000,0.4902) (5000,0.5026) (6000,0.5091) (7000,0.5134) (8000,0.5202) (9000,0.5279) (10000,0.5322) (12000,0.5429) (13000,0.5460) (15000,0.5544) (18000,0.5650) (20000,0.5733) (23000,0.5799) (26000,0.5861) (29000,0.5938) (34000,0.6005) (38000,0.6056) (44000,0.6090) (50000,0.6115) (56000,0.6137) (64000,0.6169) (73000,0.6104) (83000,0.6103) (95000,0.6113)};
\addplot[blue!60, thin, forget plot] coordinates {(1000,0.4143) (2000,0.4603) (3000,0.4923) (4000,0.5145) (5000,0.5257) (6000,0.5320) (7000,0.5393) (8000,0.5444) (9000,0.5509) (10000,0.5543) (12000,0.5606) (13000,0.5627) (15000,0.5686) (18000,0.5739) (20000,0.5797) (23000,0.5857) (26000,0.5900) (29000,0.5944) (34000,0.6031) (38000,0.6072) (44000,0.6143) (50000,0.6232) (56000,0.6358) (64000,0.6535) (73000,0.6692) (83000,0.6912) (95000,0.7048)};
\addplot[blue!60, thin, forget plot] coordinates {(1000,0.0601) (2000,0.0916) (3000,0.0957) (4000,0.1001) (5000,0.1253) (6000,0.0839) (7000,0.0757) (8000,0.0591) (9000,0.1011) (10000,0.1079) (12000,0.1140) (13000,0.1081) (15000,0.1504) (18000,0.1120) (20000,0.2061) (23000,0.1763) (26000,0.1195) (29000,0.1788) (34000,0.1928) (38000,0.1908) (44000,0.1562) (50000,0.2484) (56000,0.2183) (64000,0.3057) (73000,0.2579) (83000,0.2630) (95000,0.2386)};
\addplot[blue!60, thin, forget plot] coordinates {(1000,0.0716) (2000,0.0749) (3000,0.0724) (4000,0.0860) (5000,0.1003) (6000,0.0797) (7000,0.0817) (8000,0.0714) (9000,0.0840) (10000,0.0863) (12000,0.0873) (13000,0.0822) (15000,0.0916) (18000,0.0783) (20000,0.1216) (23000,0.1059) (26000,0.0704) (29000,0.1151) (34000,0.1262) (38000,0.1157) (44000,0.1119) (50000,0.1795) (56000,0.1613) (64000,0.2441) (73000,0.1995) (83000,0.2121) (95000,0.2056)};
\addplot[blue!60, thin, forget plot] coordinates {(1000,0.0849) (2000,0.0871) (3000,0.1004) (4000,0.1148) (5000,0.1149) (6000,0.0867) (7000,0.0805) (8000,0.0600) (9000,0.0994) (10000,0.1012) (12000,0.1160) (13000,0.1127) (15000,0.1562) (18000,0.1330) (20000,0.2187) (23000,0.2102) (26000,0.1623) (29000,0.2417) (34000,0.2634) (38000,0.2578) (44000,0.2455) (50000,0.3231) (56000,0.3253) (64000,0.3778) (73000,0.3545) (83000,0.3724) (95000,0.3851)};
\addplot[blue!60, thin, forget plot] coordinates {(1000,0.2012) (2000,0.1044) (3000,0.1064) (4000,0.1192) (5000,0.1337) (6000,0.1092) (7000,0.1091) (8000,0.0985) (9000,0.1391) (10000,0.1342) (12000,0.1460) (13000,0.1404) (15000,0.1724) (18000,0.1477) (20000,0.2163) (23000,0.2038) (26000,0.1506) (29000,0.2082) (34000,0.2153) (38000,0.2042) (44000,0.1918) (50000,0.2527) (56000,0.2454) (64000,0.2900) (73000,0.2780) (83000,0.3150) (95000,0.3418)};
\addplot[blue!60, thin, forget plot] coordinates {(1000,0.1235) (2000,0.0970) (3000,0.0926) (4000,0.1175) (5000,0.1233) (6000,0.0914) (7000,0.0821) (8000,0.0620) (9000,0.1045) (10000,0.1054) (12000,0.1131) (13000,0.1247) (15000,0.1535) (18000,0.1225) (20000,0.2340) (23000,0.1952) (26000,0.0969) (29000,0.1989) (34000,0.2108) (38000,0.1887) (44000,0.1415) (50000,0.2558) (56000,0.2399) (64000,0.3664) (73000,0.2686) (83000,0.3009) (95000,0.2749)};
\addplot[blue!60, thin, forget plot] coordinates {(1000,0.1369) (2000,0.1065) (3000,0.1082) (4000,0.1145) (5000,0.1199) (6000,0.0883) (7000,0.0758) (8000,0.0527) (9000,0.1162) (10000,0.1087) (12000,0.1229) (13000,0.1504) (15000,0.1811) (18000,0.1500) (20000,0.2660) (23000,0.2350) (26000,0.1171) (29000,0.2312) (34000,0.2471) (38000,0.2109) (44000,0.1556) (50000,0.2694) (56000,0.2525) (64000,0.3790) (73000,0.2645) (83000,0.2944) (95000,0.2612)};
\addplot[blue!60, thin, forget plot] coordinates {(1000,0.1681) (2000,0.1204) (3000,0.1177) (4000,0.1518) (5000,0.1638) (6000,0.1177) (7000,0.1016) (8000,0.0760) (9000,0.1656) (10000,0.1381) (12000,0.1470) (13000,0.1823) (15000,0.2248) (18000,0.1851) (20000,0.3284) (23000,0.3003) (26000,0.1454) (29000,0.3047) (34000,0.3134) (38000,0.2713) (44000,0.2131) (50000,0.3284) (56000,0.3134) (64000,0.4333) (73000,0.3082) (83000,0.3451) (95000,0.3031)};
\addplot[black, dashed, very thick] coordinates {(1000,0.2522) (2000,0.2030) (3000,0.1820) (4000,0.1889) (5000,0.1828) (6000,0.1399) (7000,0.1246) (8000,0.1063) (9000,0.1460) (10000,0.1371) (12000,0.1329) (13000,0.1377) (15000,0.1543) (18000,0.1227) (20000,0.1874) (23000,0.1572) (26000,0.1028) (29000,0.1371) (34000,0.1428) (38000,0.1345) (44000,0.1028) (50000,0.1676) (56000,0.1322) (64000,0.2194) (73000,0.1466) (83000,0.1495) (95000,0.1096)};
\addlegendentry{mean over all 448 heads}
\addplot[red, very thick] coordinates {(1000,0.0015) (2000,0.0026) (3000,0.0072) (4000,0.0166) (5000,0.0267) (6000,0.0417) (7000,0.0506) (8000,0.0437) (9000,0.0516) (10000,0.0491) (12000,0.0637) (13000,0.0849) (15000,0.1364) (18000,0.1989) (20000,0.2933) (23000,0.3572) (26000,0.3769) (29000,0.4491) (34000,0.5064) (38000,0.5752) (44000,0.5517) (50000,0.6546) (56000,0.6143) (64000,0.7247) (73000,0.6770) (83000,0.6693) (95000,0.6613)};
\addlegendentry{star head L1H9}
\end{axis}
\end{tikzpicture}
\vspace{-4mm}
\caption{Per-head CFAC trajectories for 760M Run A. Blue curves are the 16
heads selected by the birth census, gray dashed curves show the 10--90\%
envelope of the remaining 432 heads, and the black dashed curve is the
all-head mean. L1H9 (red) is a representative large transition, not a
coordinate that reproduces across seeds.}
\label{fig:spaghetti}
\end{figure}

Figure~\ref{fig:spaghetti} shows why mean pooling misses the effect. The blue
selected trajectories fall into two groups. One group already sits near CFAC
0.4 at the first probed checkpoint and increases steadily. A second group,
including the highlighted L1H9 (red), starts inside the gray envelope of
non-selected heads and rises above its upper edge only after its hinge
onset, saturating above 0.6. Meanwhile the gray 10--90\% band drifts
downward, and the all-head mean (black dashed) follows the majority. The
mean \emph{decreases} from 0.252 to 0.110 while the 16 selected heads rise
from 0.258 to 0.488 (non-selected 0.252 to 0.096). The mean curve alone
gives no indication that the selected subset exists. The same decomposition
appears at 1.03B Run A (mean 0.183 to 0.180, selected 0.161 to 0.313, rest
0.186 to 0.160, Appendix~\ref{app:mean}), and all nine runs are collected in
Figure~\ref{fig:spaghetti-grid} (Appendix~\ref{app:figs}).
This cancellation explains why the preregistered mean-level verdict is mixed
while the head-level census is non-empty in every run. The two analyses
measure different populations. The head-level result does not support a
model-level phase transition. It shows that the model-level average is not
an adequate statistic for sparse specialization.

\subsection{A sparse census appears across scales, but its size is not stable}

Table~\ref{tab:census} summarizes the fixed-rule birth census. On the main
analysis grids the selected population is a minority in every run, with
17/10/21 of 256 heads at 306M, 16/23 of 448 in the primary 760M runs, and
75/26 of 588 in the primary 1.03B runs (23 in Run C), a share of
3.6--12.8\% in the four primary runs. The exact size is not stable. The two
1.03B runs differ by almost a factor of three, and the 760M three-seed
16-point diagnostic triplet is 15/81/16, reaching 18.1\%.

By contrast, non-emptiness is stable across the sensitivity grid
$q\in\{0.01,0.05\}$ and $\delta_{\min}\in\{0.05,0.1,0.2\}$, and all
registered rule variants retain at least three heads per run. The evidence
therefore supports a sparse minority with substantial increases, not a
universal percentage. The 75-versus-26 gap between the two primary 1.03B
runs is the clearest instance. Many heads lie near the $\Delta_h>0.1$
boundary, so seed-level variation in how far borderline heads rise produces
large changes in census size even though sparsity, block enrichment, and
population-level rank statistics reproduce (Section~\ref{sec:positional}).
Table~\ref{tab:robust} (Appendix~\ref{app:robust}) decomposes the gate, and
Figure~\ref{fig:scale} (Appendix~\ref{app:figs}) shows census shares by
scale and run.

Onsets are also spread across training rather than synchronized. In the
onset raster (Figure~\ref{fig:raster}, Appendix~\ref{app:figs}), each 760M
run's census onsets span more than an order of magnitude, from roughly 2k
steps to past 60k, so the census identifies no single model-wide
transition, and heads join the selected population throughout training.
The five coordinates selected in both runs all lie in L0, which is
consistent with the localization result of Section~\ref{sec:positional}.

\begin{table}[h]
\centering
\caption{Birth census under the fixed main rule (BH $q=0.05$ and
$\Delta_h>0.1$).}
\label{tab:census}
\begin{tabular}{llrrr}
\toprule
Model & Run & Birth heads & \% of temporal heads & in L0 \\
\midrule
STDiT-XL/2 (760M) & Run A & 16 & 3.6\% & 8 \\
STDiT-XL/2 (760M) & Run B & 23 & 5.1\% & 8 \\
STDiT-1B/2 (1.03B) & Run A & 75 & 12.8\% & 10 \\
STDiT-1B/2 (1.03B) & Run B & 26 & 4.4\% & 12 \\
\bottomrule
\end{tabular}

\end{table}

\subsection{Reproduction is block-level, not coordinate-level}
\label{sec:positional}
\label{sec:three-seed}

The two primary 760M censuses share 5 coordinates versus 0.82 expected under
a hypergeometric null ($P=6.83\times10^{-4}$), and the two primary 1.03B
censuses share 9 versus 3.32 expected ($P=2.85\times10^{-3}$). The
initialization-controlled rank-match tests are also significant
($P=1.37\times10^{-8}$ at 760M and $P=1.45\times10^{-10}$ at 1.03B). In
the matched three-seed extension all nine within-scale pairs pass
$P<0.001$, with the 760M pairs evaluated on the diagnostic 16-point grid
(Table~\ref{tab:three-seed}, Appendix~\ref{app:stats}).

Where this signal is located determines how it should be interpreted. All
five shared 760M coordinates and six of the nine shared 1.03B coordinates
lie in L0. When L0 is excluded, the primary initialization-controlled tests
fail at 760M ($P=0.147$) and do not meet the registered 0.001 threshold at
1.03B ($P=0.02$). None of the nine pairwise tests in the three-seed
extension passes after L0 exclusion. Within L0, coordinate overlap
conditional on both runs selecting L0 heads is consistent with chance
($P=0.310/0.575$ at 760M/1.03B).

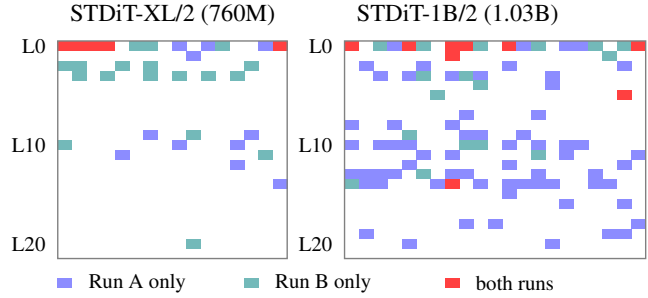
\begin{figure}[t]
\centering
\resizebox{0.62\linewidth}{!}{\begin{tikzpicture}[x=0.145cm, y=0.10cm]
\node[anchor=south west, font=\scriptsize] at (0,28.3) {STDiT-XL/2 (760M)};
\fill[red!75] (0,27) rectangle ++(1,1);
\fill[red!75] (1,27) rectangle ++(1,1);
\fill[red!75] (2,27) rectangle ++(1,1);
\fill[red!75] (3,27) rectangle ++(1,1);
\fill[teal!55] (5,27) rectangle ++(1,1);
\fill[teal!55] (6,27) rectangle ++(1,1);
\fill[blue!45] (8,27) rectangle ++(1,1);
\fill[blue!45] (10,27) rectangle ++(1,1);
\fill[teal!55] (11,27) rectangle ++(1,1);
\fill[blue!45] (14,27) rectangle ++(1,1);
\fill[red!75] (15,27) rectangle ++(1,1);
\fill[blue!45] (9,26) rectangle ++(1,1);
\fill[teal!55] (0,25) rectangle ++(1,1);
\fill[teal!55] (1,25) rectangle ++(1,1);
\fill[teal!55] (4,25) rectangle ++(1,1);
\fill[teal!55] (6,25) rectangle ++(1,1);
\fill[teal!55] (13,25) rectangle ++(1,1);
\fill[teal!55] (1,24) rectangle ++(1,1);
\fill[teal!55] (3,24) rectangle ++(1,1);
\fill[teal!55] (6,24) rectangle ++(1,1);
\fill[teal!55] (8,24) rectangle ++(1,1);
\fill[teal!55] (10,24) rectangle ++(1,1);
\fill[teal!55] (12,24) rectangle ++(1,1);
\fill[blue!45] (6,18) rectangle ++(1,1);
\fill[teal!55] (9,18) rectangle ++(1,1);
\fill[blue!45] (13,18) rectangle ++(1,1);
\fill[teal!55] (0,17) rectangle ++(1,1);
\fill[blue!45] (8,17) rectangle ++(1,1);
\fill[blue!45] (12,17) rectangle ++(1,1);
\fill[blue!45] (4,16) rectangle ++(1,1);
\fill[teal!55] (14,16) rectangle ++(1,1);
\fill[blue!45] (12,15) rectangle ++(1,1);
\fill[blue!45] (15,13) rectangle ++(1,1);
\fill[teal!55] (9,7) rectangle ++(1,1);
\draw[gray] (0,6) rectangle ++(16,22);
\node[anchor=east, font=\tiny] at (0,27.5) {L0};
\node[anchor=east, font=\tiny] at (0,17.5) {L10};
\node[anchor=east, font=\tiny] at (0,7.5) {L20};
\node[anchor=south west, font=\scriptsize] at (20,28.3) {STDiT-1B/2 (1.03B)};
\fill[red!75] (20,27) rectangle ++(1,1);
\fill[teal!55] (22,27) rectangle ++(1,1);
\fill[blue!45] (23,27) rectangle ++(1,1);
\fill[red!75] (24,27) rectangle ++(1,1);
\fill[teal!55] (25,27) rectangle ++(1,1);
\fill[red!75] (27,27) rectangle ++(1,1);
\fill[red!75] (28,27) rectangle ++(1,1);
\fill[teal!55] (29,27) rectangle ++(1,1);
\fill[red!75] (31,27) rectangle ++(1,1);
\fill[blue!45] (32,27) rectangle ++(1,1);
\fill[teal!55] (33,27) rectangle ++(1,1);
\fill[blue!45] (35,27) rectangle ++(1,1);
\fill[blue!45] (36,27) rectangle ++(1,1);
\fill[teal!55] (37,27) rectangle ++(1,1);
\fill[teal!55] (39,27) rectangle ++(1,1);
\fill[red!75] (40,27) rectangle ++(1,1);
\fill[red!75] (27,26) rectangle ++(1,1);
\fill[teal!55] (38,26) rectangle ++(1,1);
\fill[blue!45] (21,25) rectangle ++(1,1);
\fill[blue!45] (25,25) rectangle ++(1,1);
\fill[blue!45] (31,25) rectangle ++(1,1);
\fill[blue!45] (39,25) rectangle ++(1,1);
\fill[blue!45] (22,24) rectangle ++(1,1);
\fill[blue!45] (24,24) rectangle ++(1,1);
\fill[teal!55] (25,24) rectangle ++(1,1);
\fill[blue!45] (27,24) rectangle ++(1,1);
\fill[teal!55] (28,24) rectangle ++(1,1);
\fill[blue!45] (29,24) rectangle ++(1,1);
\fill[blue!45] (34,24) rectangle ++(1,1);
\fill[teal!55] (29,23) rectangle ++(1,1);
\fill[blue!45] (34,23) rectangle ++(1,1);
\fill[teal!55] (26,22) rectangle ++(1,1);
\fill[red!75] (39,22) rectangle ++(1,1);
\fill[blue!45] (28,20) rectangle ++(1,1);
\fill[blue!45] (33,20) rectangle ++(1,1);
\fill[blue!45] (20,19) rectangle ++(1,1);
\fill[blue!45] (23,19) rectangle ++(1,1);
\fill[blue!45] (27,19) rectangle ++(1,1);
\fill[teal!55] (24,18) rectangle ++(1,1);
\fill[blue!45] (28,18) rectangle ++(1,1);
\fill[blue!45] (29,18) rectangle ++(1,1);
\fill[blue!45] (32,18) rectangle ++(1,1);
\fill[blue!45] (20,17) rectangle ++(1,1);
\fill[blue!45] (22,17) rectangle ++(1,1);
\fill[blue!45] (23,17) rectangle ++(1,1);
\fill[blue!45] (24,17) rectangle ++(1,1);
\fill[teal!55] (28,17) rectangle ++(1,1);
\fill[teal!55] (29,17) rectangle ++(1,1);
\fill[blue!45] (31,17) rectangle ++(1,1);
\fill[blue!45] (33,17) rectangle ++(1,1);
\fill[blue!45] (35,17) rectangle ++(1,1);
\fill[blue!45] (36,17) rectangle ++(1,1);
\fill[blue!45] (25,16) rectangle ++(1,1);
\fill[blue!45] (27,16) rectangle ++(1,1);
\fill[teal!55] (33,16) rectangle ++(1,1);
\fill[blue!45] (35,16) rectangle ++(1,1);
\fill[blue!45] (40,16) rectangle ++(1,1);
\fill[blue!45] (24,15) rectangle ++(1,1);
\fill[blue!45] (31,15) rectangle ++(1,1);
\fill[blue!45] (39,15) rectangle ++(1,1);
\fill[blue!45] (20,14) rectangle ++(1,1);
\fill[blue!45] (21,14) rectangle ++(1,1);
\fill[blue!45] (22,14) rectangle ++(1,1);
\fill[blue!45] (23,14) rectangle ++(1,1);
\fill[teal!55] (25,14) rectangle ++(1,1);
\fill[blue!45] (27,14) rectangle ++(1,1);
\fill[blue!45] (28,14) rectangle ++(1,1);
\fill[blue!45] (33,14) rectangle ++(1,1);
\fill[blue!45] (37,14) rectangle ++(1,1);
\fill[teal!55] (20,13) rectangle ++(1,1);
\fill[blue!45] (21,13) rectangle ++(1,1);
\fill[blue!45] (22,13) rectangle ++(1,1);
\fill[blue!45] (26,13) rectangle ++(1,1);
\fill[red!75] (27,13) rectangle ++(1,1);
\fill[blue!45] (29,13) rectangle ++(1,1);
\fill[blue!45] (31,13) rectangle ++(1,1);
\fill[blue!45] (32,13) rectangle ++(1,1);
\fill[blue!45] (33,13) rectangle ++(1,1);
\fill[blue!45] (34,13) rectangle ++(1,1);
\fill[blue!45] (37,13) rectangle ++(1,1);
\fill[blue!45] (38,13) rectangle ++(1,1);
\fill[blue!45] (31,12) rectangle ++(1,1);
\fill[blue!45] (35,11) rectangle ++(1,1);
\fill[blue!45] (39,11) rectangle ++(1,1);
\fill[blue!45] (28,9) rectangle ++(1,1);
\fill[blue!45] (30,9) rectangle ++(1,1);
\fill[blue!45] (40,9) rectangle ++(1,1);
\fill[blue!45] (21,8) rectangle ++(1,1);
\fill[blue!45] (38,8) rectangle ++(1,1);
\fill[blue!45] (39,8) rectangle ++(1,1);
\fill[blue!45] (24,7) rectangle ++(1,1);
\fill[blue!45] (34,7) rectangle ++(1,1);
\draw[gray] (20,6) rectangle ++(21,22);
\node[anchor=east, font=\tiny] at (20,27.5) {L0};
\node[anchor=east, font=\tiny] at (20,17.5) {L10};
\node[anchor=east, font=\tiny] at (20,7.5) {L20};
\fill[blue!45] (0,3) rectangle ++(1,1);
\node[anchor=west, font=\tiny] at (1.3,3.5) {Run A only};
\fill[teal!55] (13,3) rectangle ++(1,1);
\node[anchor=west, font=\tiny] at (14.3,3.5) {Run B only};
\fill[red!75] (27,3) rectangle ++(1,1);
\node[anchor=west, font=\tiny] at (28.3,3.5) {both runs};
\end{tikzpicture}}
\vspace{-4mm}
\caption{Census atlas of birth-head positions by layer (rows, L0 at top) and
head (columns) for both scales. Red cells are in both runs' censuses, and
the overlap concentrates in L0.}
\label{fig:atlas}
\end{figure}

Figure~\ref{fig:atlas} shows the same result spatially. Each grid places
census heads at their (layer, head) coordinates with L0 as the top row, and
the top row is where the two runs agree. Every red (both-runs) 760M cell
and six of the nine red 1.03B cells lie in L0. Below L0, the two runs
select heads across many layers but rarely the same coordinates, which is
consistent with deeper selections being exchangeable. The supported
interpretation is therefore restricted. The first temporal block repeatedly
contains selected heads, but we find no evidence that training assigns
stable roles to the same head indices. What reproduces is the block and the
population-level ranking, not a specific coordinate such as L1H9.

\subsection{Selected heads acquire recurrent frame-attention motifs}
\label{sec:taxonomy}

The 760M frame-attention archives cover 13 of 16 Run A and 21 of 23 Run B
census heads. Every analyzed head that leaves the diffuse state ends in one
of two pattern families, self-frame diagonal or adjacent-frame band. At the
final checkpoint, Run A has 6 self-frame and 7 neighbor-band heads, and
Run B has 14 self-frame, 6 neighbor-band, and 1 still-diffuse head
(Table~\ref{tab:taxonomy}, Appendix~\ref{app:stats}). None of these 34 analyzed
census heads is classified as a single-frame sink at its final checkpoint.

The representative L1H9 transition is a diffuse-to-diagonal change, with
diagonal mass 0.062 at step 1k, 0.070 at 7k, 0.178 at 13k, 0.517 at 26k,
and 0.764 at 95k. Other selected heads instead form bands one frame behind
or ahead of the query frame, and a directional screen identifies seven
strong previous- or next-frame cases across the two runs, for example L0H11
in Run B (previous-frame mass 0.508) and L0H15 in Run A (next-frame mass
0.548). We describe these as copy-like motifs only at the level of the
attention map, since the maps do not establish contextual matching, copying
into the model output, or an induction circuit.

\begin{figure}[h]
\centering
\input{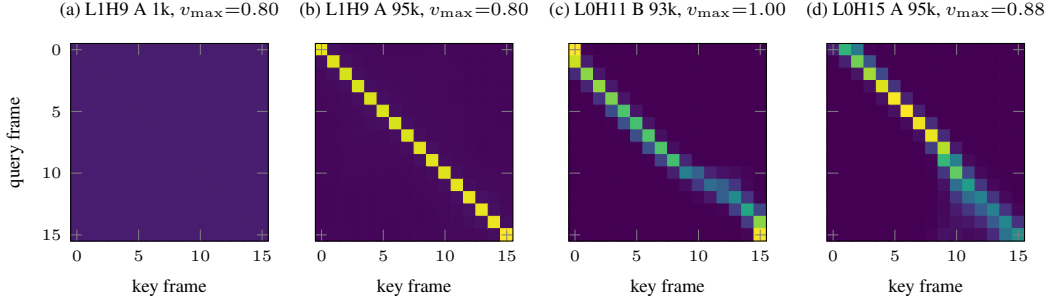}
\vspace{-7mm}
\caption{Mean temporal attention (query $\times$ key frame) of representative
census heads, rendered from the committed heatmap archives. (a) Representative
head L1H9 (Run A) at step 1000, indistinguishable from uniform (shown on the
same color scale as (b)). (b) The same head at step 95000, a self-frame
diagonal of mass $0.764$. (c) Previous-frame copy head L0H11 (Run B) at step
93000, a band one frame behind the diagonal, mass $0.508$. (d) Next-frame
head L0H15 (Run A) at step 95000, a band one frame ahead, mass $0.548$.
Color spans $0$ to the per-head maximum $v_{\max}$.}
\label{fig:patterns}
\end{figure}

Figure~\ref{fig:patterns} displays each motif family's endpoint as a
query$\times$key frame map. The checkpoint-resolved evolution of both families and the
directional statistics are collected in Figure~\ref{fig:filmstrip}
(Appendix~\ref{app:figs}) and Table~\ref{tab:directional}
(Appendix~\ref{app:stats}).

Motif families reproduce more reliably than coordinates. Self-frame,
neighbor-band, previous-frame, and next-frame patterns all occur in both
runs. The same coordinate can even reverse direction. L0H15 is
next-frame-dominant in Run A and previous-frame-dominant in Run B. What is
stable across runs is therefore the set of local temporal motifs, not a
one-to-one mapping from coordinate to function.

\subsection{Controls that bound the interpretation}
\label{sec:step0}

In the 760M probe, the L0 CFAC maximum is approximately 0.013 at step 0
versus 0.424 at step 1000 (L1H9 has 0.00053 at step 0 and 0.6613 at the
end), so the L0 pattern is absent at step 0 and present by step 1000. This
control is weak because STDiT zero-initializes the temporal-attention
output projection, so the temporal-attention branch initially contributes
nothing through that projection. The near-uniform attention observed at
step 0 is therefore an empirical property of the initialization, not a
direct consequence of the zero output projection.
\label{sec:onset}
The registered temporal-versus-spatial onset comparison is not consistent
across scales. Both 760M runs show earlier temporal onsets under both
estimators, and 1.03B Run A agrees, but 1.03B Run B shows a significant
reversal (Table~\ref{tab:onset}, Appendix~\ref{app:stats}). We therefore
do not infer a scale-general temporal-before-spatial ordering.

\begin{figure}[tbp]
\centering
\input{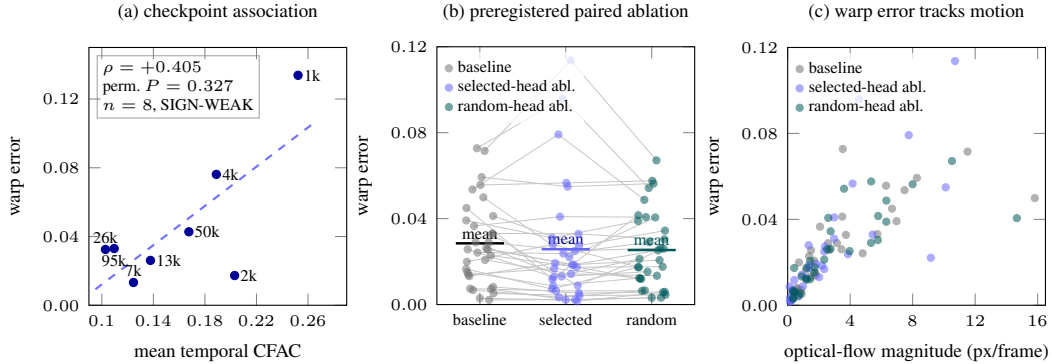}
\vspace{-9pt}
\caption{Controls that bound the interpretation. (a) The
checkpoint-level CFAC--warp-error association is sign-consistent but not
significant; $n=8$ is the number of evaluated checkpoints (1k, 2k, 4k, 7k,
13k, 26k, 50k, 95k), not a clip count. (b) The preregistered 10-head ablation is null under warp
error. (c) Warp error is strongly motion-confounded, which explains why
this metric cannot establish or refute a quality link.}
\label{fig:controls}
\end{figure}

\label{sec:d3}
The preregistered checkpoint-level association between mean temporal CFAC
and warp error is \textbf{SIGN-WEAK}, with $\rho=+0.405$ and exact
permutation $P=0.327$. The secondary selected-head series gives
$\rho=+0.191$, $P=0.665$. Warp error is also strongly coupled to motion
magnitude in the intervention data (Spearman 0.909--0.914). A preregistered
ablation of ten direction-hardened selected heads is NULL under both
registered comparisons, with $P=0.983$ (P1, born vs.\ baseline) and
$P=0.939$ (P2, born vs.\ random). An
exploratory temporal-LPIPS comparison on the same generated clips reaches
the same verdict. Both ablations reduce frame-to-frame perceptual change
together with motion, and the born-versus-random contrast is
indistinguishable. These analyses neither establish nor refute a causal
role in temporal coherence. They show that the available motion-confounded
quality metrics cannot support that claim (Appendix~\ref{app:ablation}).
Figure~\ref{fig:controls} presents these negative results. In panel~(a)
the positive rank correlation reflects the joint downward drift of both
quantities over training. In panel~(b) the three condition means (0.0285,
0.0257, 0.0254) are nearly identical. Panel~(c) shows why the metric
cannot answer the question. Within every
condition, warp error rises almost monotonically with optical-flow
magnitude, so any intervention that reduces motion also reduces warp error,
independently of temporal coherence.

\section{Discussion}

The main empirical result is a separation between population-level and
head-level dynamics. Temporal attention shows no robust mean transition, yet
a small subset of heads develops large, structured concentration increases.
This extends to video diffusion the observation that specialized heads carry
a disproportionate share of transformer
structure~\citep{voita2019heads,michel2019sixteen}. The positional result
refines this. Reproduction at the level of L0 combined with chance-level
coordinate matching within L0 suggests that the first temporal block is a
location where selected heads consistently appear, while head identities
are consistent with exchangeability at the resolution of these tests. We
find no evidence that training assigns the same role to the same index
across runs, and explaining the block-level preference would require
interventions on optimization, residual pathways, or block placement. The
motifs themselves remain hypotheses about function. Self-frame diagonals
could preserve frame-local information, and adjacent bands could route
local motion, but attention probabilities do not reveal how downstream
computation transforms the routed
information~\citep{jain2019attention,abnar2020flow}. The null ablation and
the weak quality correlation also argue against a capability claim.

We now consider how far these observations extend. Three nested
hypotheses are consistent with our evidence. The dynamics could be specific to Open-Sora,
general to factorized architectures, or a property of video DiT training in
general. We claim only the first, but the methodology is designed to test the
other two. For Latte and VDT~\citep{ma2024latte,lu2023vdt} the probe, CFAC,
and census rule apply unchanged, and for joint
architectures~\citep{wan2025wan,hacohen2024ltxvideo,yang2024cogvideox,
kong2024hunyuanvideo} marginalizing full attention over spatial positions
yields the required per-head routing distribution. If sparse,
block-localized onsets appear there, they would indicate a general feature
of video diffusion training. If they do not, the effect would depend on
factorization.

\section{Limitations}
\label{sec:notclaim}

All runs use UCF-101 and one Open-Sora STDiT family. Because the
experiments cover a single family, we cannot determine whether the observed
dynamics are Open-Sora-specific, factorized-general, or general to video
DiTs, and we accordingly position the contribution as a preregistered
methodology plus a systematic empirical case study rather than an
architecture-general claim. The analysis transfers to Latte and VDT
~\citep{ma2024latte,lu2023vdt} but is untested there, and joint
architectures require a derived routing statistic first.
Three seeds per scale support repeatability, not the full variability over
data orderings or initialization~\citep{bouthillier2021variance}, and the
scales differ in architecture, grids, and batch size, so no
result is a controlled scaling law.

The change-point null is imperfect
for log-spaced checkpoints, and head-level tests are dependent.
Benjamini--Yekutieli correction preserves the non-empty census and
positional-overlap conclusions with reduced counts
~\citep{benjamini2001dependency}, and reported BH values remain nominal
(Appendix~\ref{app:robust}). Each checkpoint uses one probe pass, and CFAC
is sensitive to the scale of attention logits. A post-hoc fixed-timestep
re-probe found the per-head $\Delta_h$ ordering stable across timesteps
(median pairwise Spearman $\rho$ 0.71--0.93 per run), but uniform logit
sharpening can raise concentration without reordering attended frames, so
scale-invariant routing statistics are needed to separate these effects.
Attention maps describe routing rather
than causal computation. The available warp metric is motion-confounded, and
the ablation is underpowered to identify subtle or compensatory functions.
No causal or perceptual-quality conclusion is supported.
Table~\ref{tab:claims} (Appendix~\ref{app:stats}) summarizes the claim
boundary.

\section{Conclusion}

We studied how temporal-attention structure arises during video diffusion
training, using a checkpoint-resolved census of every temporal head across
nine Open-Sora STDiT runs at three scales. Temporal-attention organization
is better described as sparse head-level specialization than as a
model-wide transition. A minority of heads gains substantial concentration
while population averages stay flat or decline. The reproducible positional
signal is the first temporal block rather than particular coordinates.
Attention maps among the analyzed 760M selected heads converge to a small
set of local frame-routing motifs. Correlation and ablation analyses do not
establish a quality or causal link. The preregistered methodology transfers directly to other
factorized video DiTs and, with an adapted routing statistic, to joint
architectures. Applying it there, with interventions that are robust to
motion confounds, is the natural next step.
\label{maintextend}

\FloatBarrier
\bibliographystyle{styles/iclr2027_conference}
\bibliography{references}

\subsection*{AI use statement}

Generative AI tools were used to support literature search, manuscript
restructuring, language editing, and critical review of methodological
presentation and result interpretation. They were not used to generate
experimental data or execute the reported training runs. The authors reviewed
all numerical claims, statistical conclusions, citations, and interpretations
and take responsibility for the manuscript.

\subsection*{Ethics statement}

UCF-101 is a public academic action-recognition dataset
~\citep{soomro2012ucf101}. We used it to train small research models for
analysis, not to deploy a video-generation system.
Nevertheless, video-generation methods can be misused to create deceptive or
non-consensual media. This study does not remove that risk, and downstream use
requires appropriate safeguards and compliance with the dataset's terms.

\subsection*{Reproducibility statement}

The Methods section specifies the models, checkpoint grids, read-only probe,
change-point screening gate, preregistered decision rules, and seed
comparisons. The
appendices report robustness checks and full supporting tables, and
Appendix~\ref{app:robust} records the registration timeline with the
pre-observation commit hashes. The anonymized
supplementary archive contains the preregistration documents, the metric and
probe implementations, the analysis script behind each reported result, and the
result files those scripts consume. Table~\ref{tab:compute} (Appendix~\ref{app:stats}) records the
available compute metadata. Its times are Slurm requested wall-time ceilings, not measured
elapsed times. Because per-run probe limits were not archived uniformly, the
documented minimum--maximum range is reported explicitly.

\clearpage
\appendix

\section{Why Mean-Pooled Metrics Miss Sparse Specialization}
\label{app:mean}

Table~\ref{tab:decomp} decomposes the first-to-last mean CFAC change (Run A,
both scales) into census and non-census heads. The census group rises
substantially at both scales, but it is $3.6$--$12.8\%$ of heads, and the
non-census majority is flat to falling (entropy rising) over the same
interval, so the all-head mean moves little and can even fall while the census
heads are rising. This is the arithmetic reason the mean-level D1 verdict
(MIXED) and the head-level test passing at both scales coexist without
contradiction. They measure different populations.

\begin{table}[H]
\centering
\caption{Mean CFAC at the first and last checkpoints, by head group (Run A,
with first/last = 1k/95k at 760M and 1k/50k at 1B). Computed from the committed
trajectory files.}
\label{tab:decomp}
\begin{tabular}{llrrrr}
\toprule
Model (Run A) & Head group & n & CFAC first & CFAC last & delta \\
\midrule
STDiT-XL/2 & all & 448 & 0.252 & 0.110 & -0.143 \\
STDiT-XL/2 & birth (census) & 16 & 0.258 & 0.488 & +0.230 \\
STDiT-XL/2 & non-birth & 432 & 0.252 & 0.096 & -0.156 \\
STDiT-1B/2 & all & 588 & 0.183 & 0.180 & -0.003 \\
STDiT-1B/2 & birth (census) & 75 & 0.161 & 0.313 & +0.152 \\
STDiT-1B/2 & non-birth & 513 & 0.186 & 0.160 & -0.025 \\
\bottomrule
\end{tabular}

\end{table}

\FloatBarrier
\section{Exploratory Robustness of the Census Rule}
\label{app:robust}

The 760M criteria were codified after those runs were observed, and the same
rule was fixed prospectively for the confirmatory 1B application and then applied
unchanged to the 306M and Run C extensions. The checks below are
\emph{exploratory} under the registration's change-control clause. They are
reported alongside, and do not replace, any registered verdict.

The registration timeline is as follows. The decision criteria were recorded
in \path{analysis/dense_reprobe_760m/PREREGISTRATION_D1D2D3.md} on
2026-07-08, before any 1B probe or sampling observation. The initial
registration and the pre-observation Amendment~1 correspond to commits
\texttt{35f1a12} and \texttt{07f670f}, and the file later moved in commit
\texttt{e92c8c5}, so its history requires \texttt{git log --follow}.
Because the repository is private, reviewers cannot resolve these hashes
from the submission alone, and the anonymized supplementary archive
therefore carries the referenced material directly. Post-observation
criteria changes are prohibited by the document's change-control clause,
and post-hoc tests may only be labeled exploratory.

The main rule is a conjunction of a
BH-adjusted gate and a net-rise filter $\Delta_h>0.1$, and the two components
contribute very unequally to the selection. Applied alone to the committed per-head gate files, BH at
$q=0.05$ retains a majority of heads, while $\Delta_h>0.1$ retains a small
minority, and the census is their intersection (Table~\ref{tab:robust}). The
change-point gate p-values are not established as calibrated, so the $q=0.05$
BH adjustment is treated here as nominal rather than as a formal
false-discovery-rate guarantee. The reported census is a further
data-dependent restriction of the BH-adjusted set with no separately stated
error rate. Sparsity is a property of the net-rise threshold, not of the
multiplicity correction.

Head-level tests within a run share batch
order, learning-rate schedule, and layer inputs, so BH's dependence
conditions are an assumption (Section~\ref{sec:notclaim}). Replacing BH with the
Benjamini--Yekutieli procedure, which addresses arbitrary dependence under
standard p-value validity assumptions~\citep{benjamini2001dependency},
shrinks every census but leaves
both conclusions intact. The census stays non-empty at both scales, and
cross-run positional overlap stays above chance, with $P=2.59\times10^{-4}$
at 760M (4 shared of 12/14, $0.375$ expected) and $P=3.46\times10^{-3}$ at
1B (5 shared of 33/20, $1.12$ expected), against $6.83\times10^{-4}$ and
$2.85\times10^{-3}$ under BH.

The gate's p-value is the
exceedance fraction over $20{,}000$ simulated trajectories and is not
floored, so a head whose statistic exceeds every simulated value receives
$p=0$ exactly (138 of 588 heads at 1B Run A). Recomputing every p-value with
the conventional $(r+1)/(B+1)$ estimator changes no census membership and no
overlap statistic at either scale. The censuses remain $16/23$ and $75/26$
and the overlap p-values are unchanged to the reported precision. The
resolution floor therefore affects only how the p-values are presented and
does not change any result.

\begin{table}[H]
\centering
\caption{Exploratory decomposition of the census rule and its
arbitrary-dependence variant, recomputed from the committed per-head gate
files. ``BH'' and ``BY'' are applied at $q=0.05$, and ``census'' is the
preregistered conjunction with $\Delta_h>0.1$.}
\label{tab:robust}
\begin{tabular}{lrrrrr}
\toprule
Series & heads & BH alone & $\Delta_h>0.1$ alone & census (BH $\wedge$ $\Delta_h$) & BY $\wedge$ $\Delta_h$ \\
\midrule
760M Run A & 448 & 232 & 28 & 16 & 12 \\
760M Run B & 448 & 226 & 69 & 23 & 14 \\
1B Run A & 588 & 369 & 153 & 75 & 33 \\
1B Run B & 588 & 496 & 44 & 26 & 20 \\
\bottomrule
\end{tabular}
\end{table}

\FloatBarrier
\section{Detailed Statistical Results}
\label{app:stats}

This appendix collects the full tables behind the main-text results, each
introduced by the conclusion it supports.

Table~\ref{tab:compute} records the compute allocation for the nine
training runs and the checkpoint probes. Times are requested Slurm
wall-time ceilings rather than measured elapsed times, as noted in the
reproducibility statement.

\begin{table}[H]
\centering
\small
\caption{Recorded compute allocation. All jobs used the \texttt{h100} queue on
TACC Stampede3. Times are requested Slurm wall-time ceilings, not measured
elapsed time.}
\label{tab:compute}
\begin{tabular}{@{}llcll@{}}
\toprule
Workload & Runs & Run IDs & H100 nodes/job & Time limit/job \\
\midrule
306M training & 3 & A/B/C & 2 & $\leq 26$ h \\
760M training & 3 & A/B/C & 2 & $\leq 26$--$44$ h \\
1B training & 3 & A/B/C & 4 & $\leq 22$ h \\
Checkpoint attention probes & 9 & corresponding runs & 1 &
$\leq 55$ min--$4$ h \\
\bottomrule
\end{tabular}
\end{table}

Table~\ref{tab:models} gives the exact parameter counts, block counts,
per-block temporal head counts, and checkpoint grids for the three model
configurations. The 760M three-seed common grid has 16 checkpoints, below
the 20-point minimum required for a registered main-analysis verdict, which
is why its three-seed census is reported as diagnostic.

\begin{table}[H]
\centering
\caption{Models and checkpoint grids. Parameter counts are exact. The 760M
three-seed common grid has 16 checkpoints and is diagnostic.}
\label{tab:models}
\resizebox{\textwidth}{!}{%
\begin{tabular}{lrlll}
\toprule
Architecture & Parameters (exact) & Temporal heads (blocks x heads) & Checkpoints (range) & Runs \\
\midrule
STDiT & 306,874,592 & 16 x 16 = 256 & 60 (1k-50k) & Run A / Run B / Run C \\
STDiT-XL/2 & 759,620,768 & 28 x 16 = 448 & 27 primary (1k-95k), 16 common (1k-38k) & Run A / Run B / Run C \\
STDiT-1B/2 & 1,032,795,488 & 28 x 21 = 588 & 60 common (1k-50k) & Run A / Run B / Run C \\
\bottomrule
\end{tabular}
}

\end{table}

Table~\ref{tab:three-seed} reports the matched three-scale, three-seed
reproduction. All nine within-scale initialization-controlled pairwise
tests pass $P<0.001$, and all nine fail once L0 is excluded, which is the
basis for the block-level scope of the positional claim.

\begin{table}[H]
\centering
\caption{Matched three-scale, three-seed reproduction. Census counts follow
Run A/B/C order, and each p-value column lists the three within-scale run-pair
comparisons in the order A--B, A--C, B--C.
The 760M census is diagnostic because its 16-point common grid is below the
20-point minimum required for a registered main-analysis verdict.}
\label{tab:three-seed}
\small
\setlength{\tabcolsep}{3pt}
\begin{tabular}{@{}lclll@{}}
\toprule
Scale & Grid & Census (A/B/C) & \shortstack{Init-controlled\\pairwise $P$} & \shortstack{L0-excluded\\pairwise $P$} \\
\midrule
306M & 60 (main) & 17/10/21 & 4.99e-09 / 4.22e-10 / 1.38e-09 & $>0.99$ / 0.23 / 0.136 \\
760M & 16 (diag.) & 15/81/16 & 3.42e-05 / 7.2e-12 / 3e-05 & 0.0737 / 0.058 / 0.26 \\
1B & 60 (main) & 75/26/23 & 1.45e-10 / 4.47e-09 / 1.26e-12 & 0.02 / 0.322 / 0.801 \\
\bottomrule
\end{tabular}
\end{table}

Table~\ref{tab:positional} lists the preregistered positional tests for
the primary run pairs, covering the hypergeometric census overlap, the
initialization-controlled rank-match test, and the registered L0-exclusion
scope check.

\begin{table}[H]
\centering
\caption{Positional-reproduction tests (preregistered tests (a) and (b), plus
the registered L0-exclusion scope check). Test (a) is the hypergeometric tail
probability of the census intersection over all temporal heads. Test (b)
regresses each run's per-head net increase $\Delta_h$ on an intercept and the
step-1k CFAC values of both runs, ranks the residuals, keeps the top $k$ heads
per run with $k$ equal to that run's number of heads with raw $\Delta_h>0.1$,
and applies the same hypergeometric test to the two top-$k$ sets. The
L0-excluded row repeats test (b) with the regression, ranking, and $k$
recomputed on the non-L0 heads.}
\label{tab:positional}
\resizebox{\textwidth}{!}{%
\begin{tabular}{llll}
\toprule
Model & Positional test & Overlap & $P[X \ge k]$ \\
\midrule
STDiT-XL/2 (760M) & census overlap (a) & 5 vs 0.82 expected & $P=6.83\times 10^{-4}$ \\
STDiT-XL/2 (760M) & init-controlled rank-match (b) & 17 vs 4.31 expected & $P=1.37\times 10^{-8}$ \\
STDiT-XL/2 (760M) & init-controlled, L0 excluded & 4 vs 2.1 expected & $P=0.147$ \\
STDiT-1B/2 (1.03B) & census overlap (a) & 9 vs 3.32 expected & $P=2.85\times 10^{-3}$ \\
STDiT-1B/2 (1.03B) & init-controlled rank-match (b) & 31 vs 11.45 expected & $P=1.45\times 10^{-10}$ \\
STDiT-1B/2 (1.03B) & init-controlled, L0 excluded & 12 vs 6.77 expected & $P=0.02$ \\
\bottomrule
\end{tabular}
}

\end{table}

Table~\ref{tab:onset} reports the registered temporal-versus-spatial
onset comparisons. Both 760M runs and 1.03B Run A show earlier temporal
onsets, while 1.03B Run B shows a significant reversal, which rules out a
scale-general ordering claim.
Spatial attention spans $256$ tokens per frame ($32\times32$ VAE latents
with $2\times2$ patches). The probe saves softmax probabilities for a capped
prefix of batch/frame groups in float16. The extractor renormalizes each
row, clips probabilities below $10^{-10}$, and averages row entropy over
all query tokens and retained groups before normalizing by $\log256$.

\begin{table}[H]
\centering
\caption{Onset ordering (temporal vs.\ spatial series), preregistered main
population and tests. $n$ (t/s) counts the temporal and spatial heads that
pass the registered onset population rule in each run, BH $q=0.05$ on the
change-point gate and $\Delta_h>0.05$, applied separately to the temporal
and spatial per-head series; this rule is looser than the $\Delta_h>0.1$
census, so the temporal counts differ from Table~\ref{tab:census}. MWU =
two-sided tie-corrected Mann--Whitney~\citep{mann1947test}, and perm =
20{,}000 rank-sum label permutations.}
\label{tab:onset}
\resizebox{\textwidth}{!}{%
\begin{tabular}{lllrrrrl}
\toprule
Model & Run & Estimator & n (t/s) & median onset (t/s) & MWU p & perm p & temporal first? \\
\midrule
STDiT-XL/2 & Run A & hinge & 23/145 & 10000/18000 & 0.004849 & 0.0044 & yes \\
STDiT-XL/2 & Run A & half-max & 23/145 & 18000/56000 & $4.723\times 10^{-6}$ & $5\times 10^{-5}$ & yes \\
STDiT-XL/2 & Run B & hinge & 40/80 & 10000/25000 & 0.008603 & 0.0094 & yes \\
STDiT-XL/2 & Run B & half-max & 40/80 & 6000/33000 & $1.116\times 10^{-6}$ & $5\times 10^{-5}$ & yes \\
STDiT-1B/2 & Run A & hinge & 136/92 & 10183/14185 & 0.01009 & 0.00975 & yes \\
STDiT-1B/2 & Run A & half-max & 136/92 & 19762/21116 & 0.004832 & 0.00425 & yes \\
STDiT-1B/2 & Run B & hinge & 68/152 & 9529/4910 & $4.571\times 10^{-5}$ & 0.0001 & no (reversed) \\
STDiT-1B/2 & Run B & half-max & 68/152 & 5991/3299 & $3.004\times 10^{-12}$ & $5\times 10^{-5}$ & no (reversed) \\
\bottomrule
\end{tabular}
}

\end{table}

Table~\ref{tab:taxonomy} counts the final-checkpoint motifs among the
analyzed census heads, supporting the two-family description in
Section~\ref{sec:taxonomy}.
Classification uses the attention matrix averaged over batch/spatial
positions. We compare mean self-frame, previous-frame, and next-frame mass
with the largest off-diagonal column mean (the sink score). Previous/next
means use only the 15 valid query--key pairs, without wraparound; each
column mean excludes its diagonal entry and averages over the other 15
queries. The largest feature determines the label if it is at least
$2/16=0.125$. Otherwise the label is diffuse/uniform when mean row entropy
divided by $\log16$ exceeds $0.90$, and weak/mixed otherwise. Previous- and
next-frame labels are grouped as adjacent-frame bands in the table.

\begin{table}[H]
\centering
\caption{Final-checkpoint motif counts among analyzed census heads (760M
line, where the frame-attention dumps cover temporal blocks L0--L10).}
\label{tab:taxonomy}
\begin{tabular}{lrr}
\toprule
Final-checkpoint motif & Run A (13 analyzed) & Run B (21 analyzed) \\
\midrule
self-frame (diagonal) & 6 & 14 \\
neighbor band, prev-dominant class & 5 & 5 \\
neighbor band, next-dominant class & 2 & 1 \\
still diffuse/uniform & 0 & 1 \\
\bottomrule
\end{tabular}

\end{table}

Table~\ref{tab:directional} lists the seven heads passing the directional
band-asymmetry screen, including the coordinate L0H15 whose preferred
direction reverses between runs.

\begin{table}[H]
\centering
\caption{Directional previous/next-frame-copy (induction-like) subtype,
covering census heads with final-checkpoint band asymmetry
$|\mathrm{asym}|>0.3$ and dominant $\pm1$ band mass above $0.125$ (twice
the uniform per-frame mass $1/16$). Here prev and next are the mean
attention mass on the key frame one step behind or ahead of the query frame,
$\mathrm{asym}=(\mathrm{prev}-\mathrm{next})/(\mathrm{prev}+\mathrm{next})$,
and positive values denote previous-frame dominance. This screen is applied
separately from the final-motif classification of Table~\ref{tab:taxonomy},
so a head whose largest mass is on the self-frame diagonal (final motif
``self-frame(diag)'') can still pass it.}
\label{tab:directional}
\resizebox{\textwidth}{!}{%
\begin{tabular}{llrrrll}
\toprule
Run & Head & prev band & next band & asymmetry & subtype & final motif \\
\midrule
Run A & L0H8 & 0.240 & 0.118 & +0.34 & prev-copy & prev-frame(copy) \\
Run A & L0H14 & 0.331 & 0.156 & +0.36 & prev-copy & self-frame(diag) \\
Run A & L0H15 & 0.107 & 0.548 & -0.67 & next-copy & next-frame \\
Run A & L10H12 & 0.304 & 0.138 & +0.38 & prev-copy & prev-frame(copy) \\
Run B & L0H1 & 0.127 & 0.407 & -0.53 & next-copy & self-frame(diag) \\
Run B & L0H11 & 0.508 & 0.153 & +0.54 & prev-copy & prev-frame(copy) \\
Run B & L0H15 & 0.428 & 0.080 & +0.68 & prev-copy & prev-frame(copy) \\
\bottomrule
\end{tabular}
}

\end{table}

Table~\ref{tab:claims} summarizes the claim boundary, stating for each
potential claim whether it is made and the evidence or reason.

\begin{table}[H]
\centering
\caption{Claim boundary. D1, D2, and D3 are the three preregistered
decisions (mean-level temporal emergence at 1.03B, per-head reproduction at
760M, and the metric--quality correlation). The mean-gate criterion applies
the calibrated change-point gate to the head-averaged CFAC series and
requires a significant positive break with an interior break point and a
positive post-break slope; the argmax criterion requires the series maximum
to fall after the first checkpoint; and the two-run rule requires both
primary runs to pass, otherwise the decision is MIXED.}
\label{tab:claims}
\begin{tabular}{>{\raggedright\arraybackslash}p{0.29\linewidth}>{\raggedright\arraybackslash}p{0.18\linewidth}>{\raggedright\arraybackslash}p{0.39\linewidth}}
\toprule
Claim & Status & Evidence / reason \\
\midrule
Temporal attention forms by sparse head birth & claimed & primary census 16/23 of 448 and 75/26 of 588, and 306M three-seed census 17/10/21 of 256 \\
Positional reproduction, restricted scope & claimed (L0 block + population statistics only) & all nine three-seed init-controlled pairwise tests have P<0.001, and all nine fail with L0 excluded \\
Mean-level temporal emergence & not claimed & preregistered D1 = MIXED, since 1B Run A satisfies the mean-gate criterion (U-shaped recovery) but Run B fails the argmax criterion, and the two-run rule blocks the claim \\
Temporal-before-spatial onset ordering as a scale-general law & not claimed & 1B Run B shows a significant reversal, and the 760M observation is reported as scale-local \\
Cross-run reproduction of individual head coordinates outside L0 & not claimed & none of the nine within-scale seed pairs passes P<0.001 after L0 exclusion \\
Metric-quality correlation (D3) & not claimed (SIGN-WEAK) & rho=+0.405, exact permutation P=0.327, n=8, and the CFAC-quality link is not established \\
Causal role of the born heads & not claimed (ablation NULL) & 10-head direction-hardened set, n=32, born-vs-baseline warp P=0.983, and warp error is strongly motion-confounded \\
\bottomrule
\end{tabular}
\end{table}

\FloatBarrier
\section{Additional Data-Derived Figures}
\label{app:figs}

The census atlas and the motif endpoint maps appear in the main text
(Figures~\ref{fig:atlas} and~\ref{fig:patterns}).

\begin{figure}[H]
\centering
\input{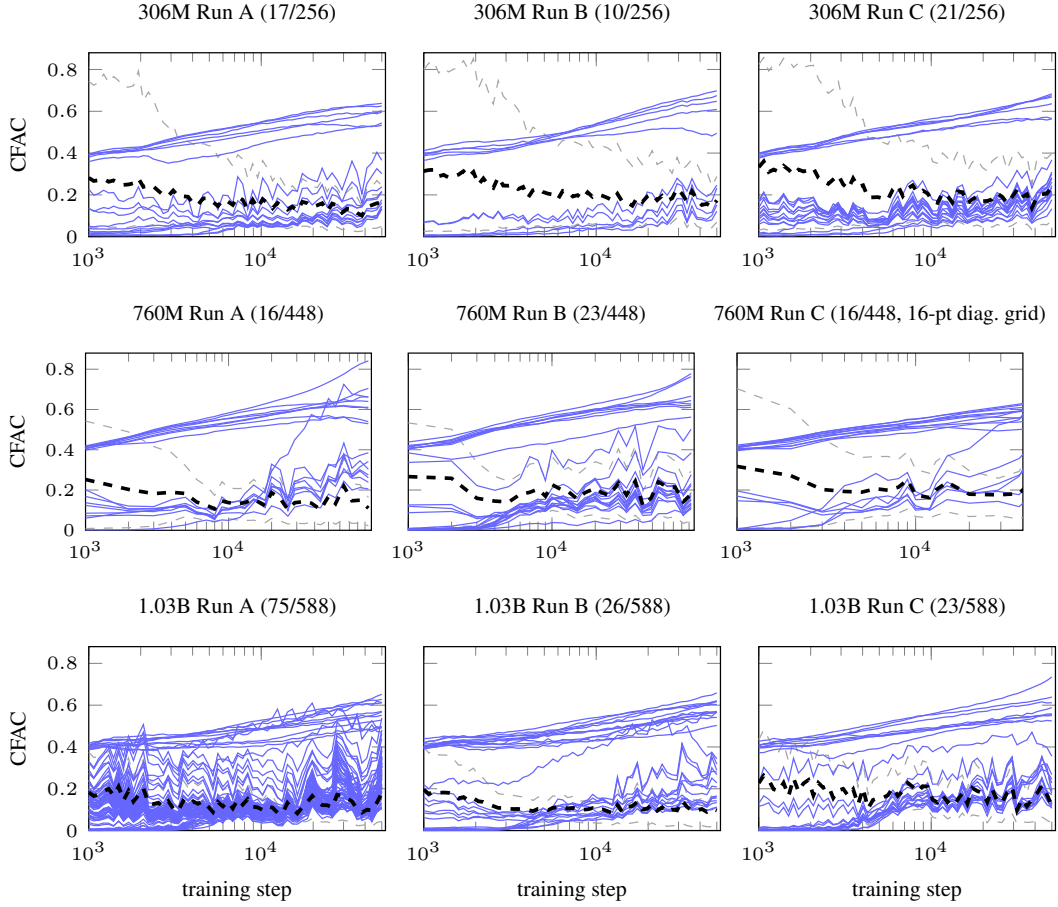}
\caption{Per-head CFAC trajectories for all nine runs. The sparse
selected-head rise against a flat or declining population reproduces in
every run. Figure~\ref{fig:spaghetti} shows the 760M Run A panel at full
size.}
\label{fig:spaghetti-grid}
\end{figure}

Figure~\ref{fig:spaghetti-grid} repeats the decomposition of
Figure~\ref{fig:spaghetti} for all nine runs, one panel per run with the
census size in each title. The same pattern appears in every panel. A
sparse set of blue selected trajectories rises above the gray 10--90\%
band while the all-head mean (dashed) stays flat or declines. The panels
also visualize the census-size instability discussed in
Section~\ref{sec:positional}. The 1.03B Run A panel shows its 75-head
census as a dense set of trajectories, many of which lie near the
effect-size boundary, whereas the neighboring 1.03B panels select sparser,
more separated sets. The 760M Run C panel is computed on the diagnostic
16-point grid and is marked as such.

\begin{figure}[H]
\centering
\resizebox{0.58\linewidth}{!}{\begin{tikzpicture}
\begin{axis}[
    ybar, width=\linewidth, height=0.5\linewidth, ymin=0,
    ylabel={birth heads (\% of temporal heads)},
    symbolic x coords={STDiT-XL/2 Run A,STDiT-XL/2 Run B,STDiT-1B/2 Run A,STDiT-1B/2 Run B},
    xtick=data, xticklabel style={font=\scriptsize},
    nodes near coords, nodes near coords style={font=\scriptsize},
    bar width=22pt]
\addplot coordinates {(STDiT-XL/2 Run A,3.57) (STDiT-XL/2 Run B,5.13) (STDiT-1B/2 Run A,12.76) (STDiT-1B/2 Run B,4.42)};
\end{axis}
\end{tikzpicture}}
\caption{Birth-census share of temporal heads in the four primary runs
(760M and 1.03B, Runs A and B). Sparsity holds in every run, while the share
varies by run and scale.}
\label{fig:scale}
\end{figure}
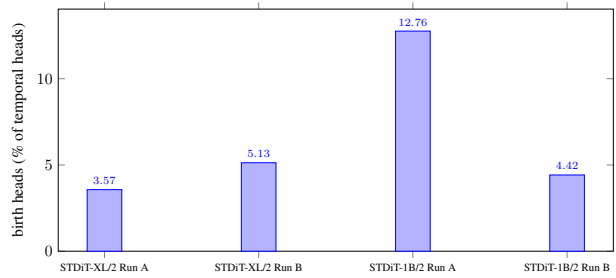

Figure~\ref{fig:scale} plots the census share of temporal heads for the
four primary runs (760M and 1.03B, Runs A and B). Sparsity holds across
scales, while the share itself varies by run and scale, which is consistent
with the census-size instability noted above.

\begin{figure}[H]
\centering
\begin{tikzpicture}
\begin{axis}[
    xmode=log, width=0.85\linewidth, height=0.42\linewidth,
    xlabel={onset step (hinge estimator)},
    xmin=1000, xmax=95000, ymin=0.3, ymax=40.7,
    enlargelimits=false, y dir=reverse,
    ytick={1,4,7,10,13,16,18,21,24,27,30,33,36,39},
    yticklabels={L00H01,L14H15,L01H09,L09H13,L00H14,L00H00,%
                 L00H15,L09H09,L03H06,L00H01,L03H10,L00H00,L02H01,L02H04},
    yticklabel style={font=\tiny}, ytick style={draw=none},
    tick label style={font=\scriptsize},
    legend pos=south west, legend style={font=\tiny, draw=none, fill=none,
        row sep=-2pt, inner sep=1pt}]
\draw[gray!35, line width=0.4pt] (axis cs:1000,16.5) -- (axis cs:95000,16.5);
\node[rotate=90, font=\scriptsize, align=center]
    at ([xshift=-2.8em]axis cs:1000,8.5) {Run A\\(16 heads)};
\node[rotate=90, font=\scriptsize, align=center]
    at ([xshift=-2.8em]axis cs:1000,29) {Run B\\(23 heads)};
\addplot[only marks, mark=*, mark size=1.4pt, blue!70] coordinates {(7000,2) (7000,3) (7000,4) (8000,5) (8000,6) (10000,7) (10000,8) (15000,9) (26000,10) (34000,12) (44000,13)};
\addlegendentry{Run A census}
\addplot[only marks, mark=*, mark size=1.4pt, teal!80!black] coordinates {(4000,20) (5000,21) (5000,22) (7000,23) (9000,24) (10000,26) (12000,28) (12000,29) (17000,30) (20000,31) (22000,32) (38000,34) (43000,35) (43000,36) (49000,37) (55000,38) (63000,39) (63000,40)};
\addlegendentry{Run B census}
\addplot[only marks, mark=*, mark size=1.4pt, red] coordinates {(6000,1) (34000,11) (50000,14) (56000,15) (64000,16) (2000,18) (4000,19) (10000,25) (12000,27) (29000,33)};
\addlegendentry{cross-run overlap}
\end{axis}
\end{tikzpicture}
\caption{Birth-onset raster (hinge estimator), STDiT-XL/2 (760M), Runs A/B.
Rows follow increasing onset time within each run, and head labels are
subsampled. Red marks denote the five shared head coordinates, all in L0.}
\label{fig:raster}
\end{figure}
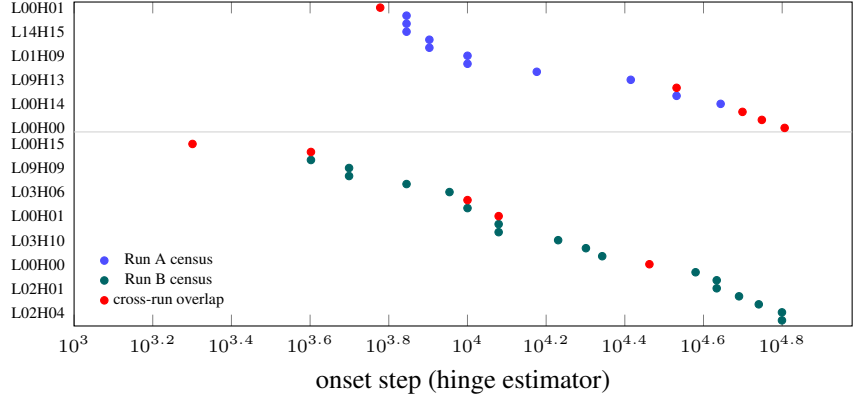

Figure~\ref{fig:raster} shows the onset raster behind the timing discussion
in the main text. Each row is one 760M census head and each mark its hinge
onset on a logarithmic axis. Onsets span roughly 2k to past 60k steps in
both runs, and the red rows, the five cross-run overlap coordinates, all
carry L0 indices.

\begin{figure}[H]
\centering
\input{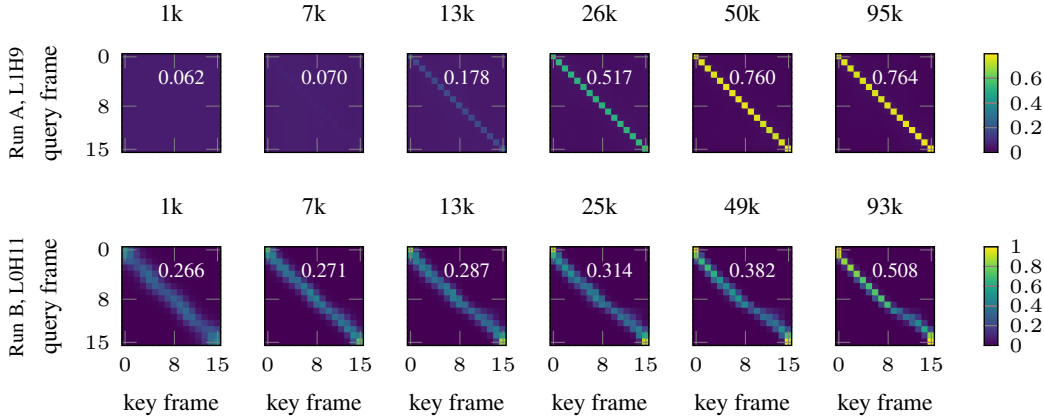}
\caption{Checkpoint-resolved evolution of two representative census heads.
In the top row, Run A L1H9 transitions from diffuse attention to a
self-frame diagonal (diagonal mass 0.062 at 1k to 0.764 at 95k). In the bottom row, Run B L0H11 sharpens from a broad self-frame band toward
a previous-frame copy band (previous-frame mass 0.266 at 1k to 0.508 at 93k).
The relevant mass is printed in each panel.}
\label{fig:filmstrip}
\end{figure}

The clearest individual trajectory is 760M Run A head L1H9 (CFAC 0.00147 at
1k to 0.6613 at 95k, peak 0.725, hinge onset near 10k, saturation after
50k). We treat it only as an interpretable example of the selected class.
Figure~\ref{fig:filmstrip} adds the time axis that the endpoint maps of
Figure~\ref{fig:patterns} omit. In the top row, L1H9 is flat at steps 1k
and 7k (diagonal mass 0.062 and 0.070, at the uniform baseline), a faint
diagonal becomes visible at 13k (0.178), dominates the map by 26k (0.517),
and is saturated by 50k (0.760) with essentially no further change to 95k
(0.764). This is an abrupt transition. In the bottom row, L0H11 instead starts from
an already-broad self-frame band at 1k (previous-frame mass 0.266) and
sharpens gradually and monotonically toward the previous-frame copy motif
(0.287 at 13k, 0.382 at 49k, 0.508 at 93k). The two motif families thus
reach their final forms on different schedules. One changes abruptly and
the other sharpens steadily.

\FloatBarrier
\section{Causal Ablation of the Born Heads, a Preregistered Null with a
Metric Confound}
\label{app:ablation}

The main results are descriptive. To test a causal prediction, we
preregistered an intervention before video generation
(\texttt{analysis/ablation/PREREG\_HEADS.md}). The \emph{born} intervention
uses the 10-head direction-hardened Run A set (BH $q=0.05$,
$\Delta_h\ge0.1$, direction $=+1$), not the 16-head primary census. Its
coordinates are L0H10, L0H15, L1H9, L9H6, L9H13, L10H8, L10H12, L11H4, L12H12,
and L14H15. The random control uses 10 disjoint non-census heads drawn with
fixed seed 20260718: L2H15, L5H10, L12H13, L13H7, L13H15, L14H7, L19H4, L22H8,
L23H0, and L25H10. Baseline, born-ablated, and random-ablated conditions generate the
same 32 prompts with identical sampled noise. Ablation zeroes selected
per-head outputs before the temporal-attention output projection
~\citep{michel2019sixteen,voita2019heads}.

The primary one-sided paired tests
ask whether born-head ablation raises warp error relative to baseline (P1)
and random-head ablation (P2). Neither approaches significance.
Mean warp error is $0.0285$ (baseline), $0.0257$
(born-ablated), and $0.0254$ (random-ablated), with
$P=0.983$ for P1 and $P=0.939$ for P2. Under the fixed decision
grid the verdict is NULL, and at $n=32$ the experiment does not detect any
degradation from born-head ablation. Random ablation instead lowers warp error
relative to baseline (report-only two-sided $P=0.00022$).
Table~\ref{tab:ablation} reports the full paired results.

As a post-hoc motion check, mean
optical-flow magnitude is
$3.549$ px/frame at baseline, $2.654$ after
born-head ablation ($P=0.00549$), and
$3.227$ after random-head ablation
($P=0.00409$). The born-vs-random motion contrast is not
significant ($P=0.062$). Within each condition, motion
magnitude and warp error are almost monotone (Spearman
$0.909$--$0.914$). Warp error therefore decreases whenever motion decreases, so it is
confounded for this intervention.

As a post-hoc,
non-preregistered complement to the warp metric, we compute per-clip temporal
LPIPS~\citep{zhang2018perceptual} (mean AlexNet LPIPS between consecutive
frames, \texttt{analysis/ablation/temporal\_lpips.py}) on the same 96 stored
clips. Means are $0.140$ (baseline), $0.119$ (born-ablated), and $0.127$
(random-ablated). Paired two-sided sign-flip permutation tests ($n=32$,
$200{,}000$ resamples) give $P=0.027$ for born-ablated versus baseline,
$P=0.00002$ for random-ablated versus baseline, and $P=0.378$ for born-ablated
versus random-ablated. Both ablations reduce frame-to-frame perceptual change
relative to baseline, consistent with the motion reduction above, and the
born-versus-random contrast remains indistinguishable. This exploratory check
does not alter the preregistered NULL.

We retain the preregistered NULL, make no causal
claim in either direction, and do not claim that the born heads uniquely
drive motion. The result explains why D3 and the intervention cannot
establish a perceptual-quality link with the current metric. It does not
alter the descriptive birth, block-enrichment, or motif results.

\begin{table}[H]
\centering
\caption{Head-ablation results over 32 paired clips. P1 (born vs.\ baseline)
and P2 (born vs.\ random) are the preregistered one-sided paired tests of
whether ablation raises warp error. The random-vs.-baseline warp comparison is
report-only and two-sided, and the motion analyses are post hoc and two-sided.}
\label{tab:ablation}
\resizebox{\textwidth}{!}{%
\begin{tabular}{lrrrrr}
\toprule
Condition & Mean warp & $p$ vs.\ base (warp) & $p$ vs.\ random (warp) & Mean $|\mathrm{flow}|$ & $p$ vs.\ base (motion) \\
\midrule
baseline & $0.0285$ & n/a & n/a & $3.549$ & n/a \\
born-ablated & $0.0257$ & $0.983$ (P1, one-sided) & $0.939$ (P2, one-sided) & $2.654$ & $0.00549$ \\
random-ablated & $0.0254$ & $0.00022$ (two-sided) & n/a & $3.227$ & $0.00409$ \\
\bottomrule
\end{tabular}%
}

\end{table}

\FloatBarrier

\end{document}